\documentclass[a4paper]{article}
\pdfoutput=1

\usepackage{lpic}
\usepackage{INTERSPEECH2021}
\usepackage{multirow}
\usepackage{graphicx}
\usepackage{subcaption}
\usepackage{microtype}
\usepackage{amsmath,amssymb}
\usepackage{nicefrac}
\usepackage{cite}

\title{\vspace{-1mm}On the limit of English conversational speech recognition}
\name{\vspace{-6.5mm}\\Zolt{\'a}n T{\"u}ske, George Saon, Brian Kingsbury}
\address{\vspace{-1.5mm}IBM Research AI, Yorktown Heights, USA}
\email{zoltan.tuske@ibm.com}

\begin{document}
\ninept

\renewcommand{\baselinestretch}{0.938}\normalsize
\newcommand{\argmax}{\operatornamewithlimits{argmax}}
\maketitle

\begin{abstract}
\vspace{-1mm}
In our previous work we demonstrated that a single headed attention encoder-decoder model is able to reach state-of-the-art results in conversational speech recognition.
In this paper, we further improve the results for both Switchboard 300 and 2000.
Through use of an improved optimizer, speaker vector embeddings, and alternative speech representations we reduce the recognition errors of our LSTM system on Switchboard-300 by 4\% relative.
Compensation of the decoder model with the probability ratio approach allows more efficient integration of an external language model, and we report 5.9\% and 11.5\% WER on the SWB and CHM parts of Hub5'00 with very simple LSTM models.
Our study also considers the recently proposed conformer, and more advanced self-attention based language models.
Overall, the conformer shows similar performance to the LSTM; nevertheless, their combination and decoding with an improved LM reaches a new record on Switchboard-300, 5.0\% and 10.0\% WER on SWB and CHM.
Our findings are also confirmed on Switchboard-2000, and a new state of the art is reported, practically reaching the limit of the benchmark.
\end{abstract}

\vspace{-.25mm}
\section{Introduction}
\vspace{-.5mm}
\label{sec:intro}
End-to-end (E2E) automatic speech recognition (ASR) models directly map an acoustic feature sequence to a word sequence.
Due to their universal capability to handle even non-monotonic alignment, attention models are widely used in many machine learning problems, e.g. translation \cite{Bahdanau2015}.
The extreme flexibility of these models also allows efficient transfer learning to a non-monotonic spoken language understanding problem from a monotonic speech recognition task \cite{Kuo2020}.
Furthermore, the attention model also holds the best record on many ASR tasks, e.g. \cite{Gulati2020}.
In \cite{Tuske2020}, we have shown that attention models are able to reach state-of-the-art recognition performance even if only 300 hours of speech are available, using a long list of various regularization techniques.
This work extends the best recipe proposed in that study to further advance recognition quality.
The overview of the investigated methods and related works is presented in Section~\ref{sec:meth}.
The details about the updated experimental settings and results are presented in Section~\ref{sec:exps} and~\ref{sec:expr}.
In Section~\ref{sec:conc}, the paper closes with conclusions.

\vspace{-1mm}
\section{Methods}
\vspace{-.0mm}
\label{sec:meth}
Extending our previous study in \cite{Tuske2020}, by default we use various dropout methods, data augmentation and regularization approaches, curriculum learning and scheduled sampling techniques 
to mitigate the inherent data sparsity problem of direct sequence-to-sequence modeling \cite{hinton2012,pmlr-v28-wan13,
Kanda2013,Ko15,Saon2019,Park2019,
Murray1994,Krueger2017,Szegedy2016,Krogh1992,
Bengio2009,Bengio2015,Amodei2016}.
On top of those, the following methods and modeling approaches are investigated in this paper.
\\ {\bf AdamW optimizer}:
The Adam optimizer combines adaptive stochastic gradient descent (SGD) optimization and classic momentum \cite{kingma2014adam}.
Recently, a modification has been proposed, in which weight decay is decoupled from the adaptive gradient calculation \cite{loshchilov2018decoupled}.
\\ {\bf I-vector} based speaker adaptation estimates an identity vector from incoming acoustic features.
This vector transforms the parameters of a universal background model (UBM) in a shared sub-space to match the speaker dependent distribution.
I-vectors have been successfully used for neural network based ASR, e.g. in \cite{saon13,Kitza2019,saon21}.
To combine with speed and tempo perturbation, the artificially created recordings of a speaker are treated as coming from a previously unseen speaker.
\\ {\bf Alternative log-Mel representations}:
in the standard feature extraction pipeline, amplitude spectra are usually extracted every 10ms, applying a 25-32ms time window.
Studies on acoustic modeling of raw waveforms, however, suggest that higher temporal resolution might be beneficial \cite{Zhu2016,tuske2018:waveform}.
Therefore, we implement and investigate a reparametrized log-Mel extraction pipeline in which
amplitude spectra are extracted every 2.5ms using a 10ms analysis window.
Further, 7th root compression is applied instead of logarithmic, as in Gammatone or PLP features \cite{Hermansky1990,schlueter07}.
The alternative feature extraction pipeline is also extended with small-energy masking (SEM) perturbation \cite{Kim2020}.
Relative to the peak energy in a given utterance, the method masks time-frequency bins with small energy in the Mel-spectral domain.
We also propose to apply high energy clipping (HEC) distortion on the Mel-spectrum, on top of SEM.
Denoting the amplitude spectrum as $e_{t,c}$, where $t$ corresponds to the frame index and $c$ the filterbank channel, the following function is applied:

\vspace{-5.9mm}
\begin{align}
\eightpt
 e^{(\text{HEC})}_{t,c} = \min(\eta_{c}, e_{t,c})
\end{align}
\vspace{-3.5mm}

\noindent where $\eta_{c}$ denotes a randomly chosen, channel dependent threshold above which the amplitude values get clipped, e.g. picking the 80th percentile per channel.
\\ {\bf Convolutional self-attention modules:}
Recurrent hidden layers, e.g. long short-term memory (LSTM), are often chosen to model temporal dependencies in the speech signal \cite{Hochreiter97}.
Self-attention based transformer models have been explored as encoder and also as decoder  for ASR in \cite{Sperber2018,Povey2018,Zhang2020,Dong2018}.
Recently, the combination of convolutional neural network and transformer has been proposed and achieved state-of-the-art recognition results on read speech \cite{Gulati2020}.
Our study also investigates the effect of replacing the LSTM based encoder, as well as the replacement of our single-head LSTM decoder with a conformer.
\\ {\bf Self-attention based language models}
can easily capture long term dependencies and can outperform LSTM models \cite{AlRfou}.
Further improvement was made to the model by introducing relative positional encoding and increased context length through segment-level recurrence \cite{dai2019}.
As shown by \cite{Tuske2018,Xiong2018}, significant gain has been observed on the Switchboard test sets when the language model (LM) is trained at the conversation level.
In this paper, we also investigate whether a cross sentence transformer or conformer LM outperforms the LSTM LM, similar to \cite{irie19:asru}.
\\ {\bf External language model fusion through probability ratio:}
as proposed in \cite{McDermott2019}, to decode with an end-to-end sequence posterior $p^{(\text{\tiny E2E})}(w_1^N|x_1^T)$ and external language model $p^{(\text{\tiny EXT})}(w_1^N)$, the language model learned by the E2E model $p^{(\text{\tiny E2E})}(w_1^N)$  has to be compensated.

\vspace{-4mm}
\begin{align}
\eightpt
 p(w_1^N|x^T_1) \sim \frac{p^{(\text{\tiny E2E})}(w_1^N|x^T_1)}{p^{(\text{\tiny E2E})}(w_1^N)} \cdot p^{(\text{\tiny EXT})}(w_1^N) \label{eq:probratio}
\end{align}
\vspace{-3mm}

\noindent where the fraction is proportional to $p(x^T_1|w_1^N)$, since $p(x_1^N)$ can be ignored during search.
Thus, Eq.~\ref{eq:probratio} is basically the famous hybrid model equation applied at the sequence level \cite{Bourlard1993}.
Unlike the hybrid approach where the framewise priors can directly be estimated and integrated into the softmax layer, either using an alignment or by approximate marginalization using the training samples and the posterior distribution \cite{Manohar2015}, it is difficult to estimate the sequence prior $p^{(\text{\tiny E2E})}(w_1^N)$.
The usual solution is to estimate an additional, e.g. LSTM, model on the transcription of acoustic data and plug it into Eq.~\ref{eq:probratio} after position- and sequence-level smoothing, according to the next paragraph.
\\ {\bf Model combination:}
an encoder-decoder model provides position-wise normalized scores $p(w_n|w_1^{n-1},x_1^T)$; thus, it is easy to apply classifier combination rules that use multiple models in making a decision \cite{Kittler1998,Kirchhoff2000}.
Log-linear score combination and joint decoding with various models is an old idea in ASR \cite{Dugast1994}, and shallow fusion simply corresponds to sequence-level log-linear combination \cite{GulcehreFXCBLBS15}.
Extending the concept to $K$ LM and $L$ E2E models, in this paper the following decision rule is used to find the optimal sequence:

\vspace{-5mm}
\begin{align}
  \eightpt
  \hat{w}_1^{\hat{N}} = \argmax\limits_{w_1^N,N} \left( \sum_{n=1}^{N} \sum_{k=1}^{K}\lambda_{\text{LM}_k} \log p_k(w_n|w_1^{n-1}) \right . \nonumber \\[-0.1em]
   + \left. \sum_{n=1}^{N} \sum_{l=1}^{L} \lambda_{\text{E2E}_l} \log\ p_l(w_n|w_1^{n-1},x_1^T)  \right. \label{eq:combo} \\[-0.1em]
  + \left. \sum_{l=1}^{L} \lambda_{\text{CT}_l} \sum_{t=1}^{T'} I_{\{ \alpha > \tau_l \}} (\max_n \alpha_{l,n,t}) +  \lambda_{\text{LEN}}N \right) \nonumber
\end{align}
\vspace{-2mm}

\noindent where the first term corresponds to sequence level log-linear interpolation of multiple external language models, and includes the probability ratio model with negative weight.
The second term corresponds to the combination of the end-to-end ASR models.
The third term (applied only to single-head decoder) is the coverage term (CT), using decoder specific threshold $\tau_l$ the indicator function $I$ gives extra credit to hypotheses which cover the subsampled acoustic feature stream ($T\rightarrow T'$) with sharp decoder attention $\alpha_{n,t}$ \cite{ChorowskiJ16,Tu2016}.
The last term controls the reward for emitting longer hypotheses \cite{Hannun14}.
In addition, the position-wise priors and posteriors are also smoothed: $\tilde{p}_m(w_n|\cdot) = p_m(w_n|\cdot)^{\beta_m} / \sum_{w_n} p_m(w_n|\cdot)^{\beta_m}$.

\section{Experimental setup}
\label{sec:exps}
Our research focuses on the standard Switchboard (SWB) English conversational telephony speech recognition benchmark.
Data preparation follows the Kaldi \texttt{s5c} recipe \cite{Povey_ASRU2011} and the work of \cite{Saon2017}.
The setup is based on our previous best system; for further details refer to \cite{Tuske2020}.
Below, we focus mainly on the settings of the new components.

\vspace{-1.5mm}
\subsection{Input features}
\vspace{-1mm}
Unless noted otherwise, 80-dimensional log-Mel features are extracted from a 25ms long window shifted by 10ms.
When using the 2.5ms frame rate, the window size is reduced to 10ms, the number of Mel filters to 20, and the SpecAugment parameters are also adjusted, e.g. the temporal width of the mask is quadrupled.
Results with SEM and HEC are only presented for high temporal resolution log-Mel features; standard features showed marginal improvement together with SpecAugment.
SEM is turned on with 10\% probability, the peak energy corresponds to the 95th percentile and the energy threshold is chosen randomly between -30 and -20dB.
Instead of masking after normalization, we reset the Mel-filter outputs to the channel-wise utterance-level mean values.
The high energy clipping is applied 40\% of the time, and the channel-wise threshold $\eta_{c}$ is randomly selected from the interval of 80-100th percentile.
The HEC operation is followed by rescaling to preserve the total energy of an utterance.
Two frames of the high frame rate features are stacked and every second frame is skipped.
The 100-dimensional i-vectors are extracted using a 2048-component 40-dimensional diagonal covariance Gaussian mixture UBM trained on PLP features transformed with LDA and semi-tied covariance transform.

\vspace{-1.5mm}
\subsection{Sequence-to-sequence models}
\vspace{-1mm}
The 300 and 2k-hour systems are built on graphemic units created by 600 and 1000 BPE rules \cite{subword-nmt}.
Our default E2E model has a 6-layer bidirectional LSTM encoder and 2-layer unidirectional LSTM decoder.
The conformer based decoder follows the structure of \cite{Vaswani2017}.
The second multi-head attention units, which use absolute positional encoding and the encoded sequence as key and value, are inserted after the convolutional block.
We note that our LSTM decoder has a single-head location-aware additive attention mechanism, while the conformer decoder has multiple layers of multiplicative multi-head attention \cite{chorowski15,Gulati2020}.
In the SWB-300 experiments, the number of decoder and encoder conformer layers is limited to 3 and 12.
The model and the inner dimensions of the feed-forward and convolution modules are set to 384, 1536, and 1280, and the number of heads and head dimension are 6 and 96.
In the 2000-hour experiments 16-layer encoder and 6-layer decoder conformer modules are used, and the head and inner dimensions are increased to 128 and 2048.
The dropout rates in the conformer modules are set to 15\% and 20\% for the 300 and 2000-hour setups.
The input to the conformer encoder is processed by two 2D convolutional units with 128 and 256 output channels using 5x5 kernels, ReLU non-linearity and 10\% dropout.
The kernels are strided by two in each direction.
Similarly, the LSTM encoder reduces the frame rate by a factor of 4 (or 8) to 1/40ms by pyramidal processing \cite{Chan2016}.
The i-vectors are concatenated either to the inputs of every LSTM encoder layer or position-wise feed-forward layer of a conformer.
In order to use a highly parallelized implementation, zoneout and scheduled sampling are not applied to the conformer decoder.
In each model, the batch normalization layers are frozen and turned into global normalization layers in the middle of the training \cite{Tuske2020}.
The SWB-300 models are trained on 6 V100 GPUs with a batch size of 32 sequences per GPU.
The SWB-2000 models are optimized on 24 GPUs with variable batch size of up to 128 sequences per GPU; processing 2000 hours of speech e.g. by conformer-conformer model took about 20 minutes.
In every case, the first iterations are used to warm up the learning rate \cite{Goyal2017}.
The learning rate is then kept constant and annealed exponentially in the last 25\% of the training by a total factor of 256.
When switching from SGD to AdamW, the initial learning rate had to be reduced by a factor of 30.
Overall 450-500k updates are performed until convergence to maximize sentence posterior probability (cross-entropy training).

\vspace{-1mm}
\subsection{Language models}
\vspace{-1mm}
The language models are trained on the 24M-word Switchboard+Fisher data.
As in our previous study, the unidirectional LSTM language model has two layers and 2048 nodes per layer.
The transformer-XL language model is based on the implementation of \cite{nvidiaXl,dai2019},
having 10 self-attention layers, 8 heads with 64 dimension per head, and the inner and model dimension is set to 2048 and 512.
The conformer LM has a similar structure, but the inner dimension in the feed-forward blocks is reduced to 1536 and 768, and the convolution module's inner dimension is set to 768, and batch normalization is deactivated.
The macaron-like, second feed-forward units are also switched off.
For both models, the dropout rates are set to 0.05, and the memory of the self-attention is unconstrained, limited only by the input sequence length.
The cross-utterance models are initialized with the model trained on independent utterances.
Their input is constructed by concatenation of successive utterances of a conversation channel up to 150 words.
The denominator model of Eq.~\ref{eq:probratio} follows the decoder structure of a corresponding E2E model, and is trained on the reference transcriptions of the acoustic data.
All models are trained by SGD optimization with batch size of 256 sequences and Nesterov momentum \cite{nesterov1983}.

The decoding hyper parameters are optimized on Hub5'00, iteratively by one dimensional grid search.
During search, the beam size is limited to 16.
To validate our findings and avoid learning on the test, we also evaluated the best systems on RT03 and Hub5'01, measuring word error rate (WER).

\begin{table}
  \eightpt
  \centering
  \caption{Effect of AdamW optimizer and i-vector on LSTM based encoder-decoder with and without using external language model in shallow fusion. Measured on Switchboard-300.}
  \vspace{-3.5mm}
  \begin{tabular}{|@{}c@{}|@{}c@{}||c|c|c||c|c|c|}
    \hline
\multirow{2}{*}{\hspace{1mm}optimizer\hspace{1mm}} & \multirow{2}{*}{\hspace{1mm}ivec.\hspace{1mm}}     & \multicolumn{3}{c||}{w/o LM} & \multicolumn{3}{|c|}{w/ LM} \\
\cline{3-8}
                           &                            &  swb & chm & tot. & swb & chm & tot. \\
\hline            
\hline            
              SGD       &                               &  7.5 & 14.8 & 11.2 & 6.4 & 13.2 & 9.8  \\ \cline{1-1}\cline{3-8}
\multirow{2}{*}{AdamW}  &                               &  6.9 & 13.8 & 10.4 & 6.1 & 12.6 & 9.4  \\ \cline{2-8}
                        & \multirow{1}{*}{$\checkmark$} &  \bf{6.8} & \bf{13.5} & \bf{10.2} & \bf{6.0} & \bf{12.4} & \bf{9.3}  \\
\hline
\end{tabular}                              
\label{tab:lstm_adamw_ivec}
\vspace{-2mm}
\end{table}

\begin{table}
  \eightpt
  \centering
  \caption{Experiments with alternative log-Mel speech representation on Switchboard-300 using LSTM model. WER measured after decoding with cross-utterance LM in shallow fusion.}
  \vspace{-3.5mm}
  \begin{tabular}{|c|c|@{}c@{}|@{}c@{}|@{}c@{}|@{}c@{}||c|c|c|}
    \hline
\multicolumn{3}{|c|}{logMel}             &\multirow{3}{*}{\parbox[c]{6mm}{SEM\\HEC}}&  \multirow{3}{*}{\hspace{2mm}optim.\hspace{2mm}} & \multirow{3}{*}{\hspace{0.8mm}ivec.\hspace{0.8mm}} & \multicolumn{3}{c|}{WER [\%]} \\ \cline{7-9} \cline{1-3}
\multicolumn{2}{|c|}{win. [ms]}                 & \multirow{2}{*}{dim.} &      &              &        &  \multicolumn{3}{c|}{hub5'00} \\ \cline{1-2}\cline{7-9}
     step      &  size      &        & \hspace{7mm} & \hspace{2mm} &        &   swb   &  chm   & tot. \\
\hline
\hline
     10        &   25      & \hspace{1.7mm}80\hspace{1.7mm} &                                &  AdamW         &  $\checkmark$                 & {\bf 6.0} & \bf{12.4} & \bf{9.3} \\
\hline
\hline
\multirow{4}{*}{2.5} & \multirow{4}{*}{10} & \multirow{4}{*}{20} &                                & \multirow{2}{*}{SGD}           &                               & 6.5     & 13.1   & 9.8  \\ \cline{4-4} \cline{7-9}
               &           &             &  \multirow{3}{*}{$\checkmark$} &              &                               & 6.2     & 13.0   & 9.6  \\ \cline{5-5}\cline{7-9}
               &           &             &                                & \multirow{2}{*}{AdamW}        &                               & 6.2     & 12.3   & 9.3  \\ \cline{6-6}\cline{7-9}
               &           &             &                                &              & \multirow{1}{*}{$\checkmark$} & \bf 6.2 & \bf 12.0 & \bf 9.1  \\
\hline
\end{tabular}                              
\label{tab:hires}
\vspace{-4mm}
\end{table}

\vspace{0mm}
\section{Experimental results}
\vspace{0mm}
\label{sec:expr}
\subsection{Comparison of SGD and AdamW optimizers, effect of i-vector on LSTM model}
\vspace{-1mm}
As can be seen in Table~\ref{tab:lstm_adamw_ivec}, AdamW significantly improves the quality of the LSTM model, on average by 7\% without LM and 4\% after shallow fusion with cross-utterance LM.
I-vectors improve the results further, and give small but consistent gain after decoding with LM.
We note that the best result is produced with a 57M-parameter E2E model in Table~\ref{tab:lstm_adamw_ivec}, and it already outperforms the 280M-parameter model developed in \cite{Tuske2020}.
In Table~\ref{tab:hires}, we test our alternative, high temporal resolution log-Mel speech representations.
Comparing to the best results in Table~\ref{tab:lstm_adamw_ivec}, it can be observed that the two features perform similarly.
The systems turned out to be complementary, and their combination resulted in significant gain, see e.g. Table~\ref{tab:lm2}.

\begin{table}
  \eightpt
  \centering
  \caption{Performance comparison of conformer and LSTM encoder/decoder blocks on Switchboard-300. External LSTM LM operates across utterances using shallow fusion.}
  \vspace{-3.5mm}
  \begin{tabular}{|@{}c@{}|@{}c@{}|@{}c@{}|@{}c@{}|@{}c@{}|@{}c@{}|c|c|@{}c@{}|}
    \hline
 \multicolumn{2}{|c|}{model} & \multirow{2}{*}{optim.} &  sched. & \multirow{2}{*}{ivec.} &  ext.        & \multicolumn{3}{c|}{hub5'00}  \\ \cline{1-2}\cline{7-9}
 enc.    &    dec.   &           &  samp.  &              &  LM          &     swb    &    chm     & \hspace{2mm}tot.\hspace{.5mm} \\
\hline
\hline
\multirow{1}{*}{LSTM}& \multirow{1}{*}{LSTM} &   
         AdamW   & $\checkmark$ & $\checkmark$ & $\checkmark$ &   6.0      & 12.4       & \textcolor{white}{0}9.3 \\
\hline
\hline
\multirow{5}{*}{Conf.}   & \multirow{5}{*}{LSTM}   & \multirow{1}{*}{SGD} & \multirow{5}{*}{$\checkmark$} & \hspace{7mm} & \hspace{6mm} &   6.8      & 13.5       & 10.2 \\ \cline{3-3}\cline{7-9}
         &           & \multirow{4}{*}{AdamW} &              &              &              &   6.7      & 13.0       & \textcolor{white}{0}9.8  \\ \cline{6-9}
\hspace{10mm} & \hspace{10mm} & \hspace{12mm} & \hspace{9mm} &              & $\checkmark$ &  \bf 5.8      & 12.0       & \bf \textcolor{white}{0}8.9  \\ \cline{5-5}\cline{6-9}
         &           &           &              & \multirow{2}{*}{$\checkmark$} &              &   6.5      & 12.9       & \textcolor{white}{0}9.7  \\ \cline{6-9}
         &           &           &              &              & $\checkmark$ & 5.9      & \bf 11.8       & \bf \textcolor{white}{0}8.9  \\
\hline
\hline
\multirow{6}{*}{Conf.} & \multirow{6}{*}{Conf.} &   SGD     &              &              &              &   6.9      &  14.2      & 10.5 \\ \cline{3-3}\cline{7-9}
         &           & \multirow{5}{*}{AdamW} &              &              &              &   7.0      &  13.6      & 10.3 \\ \cline{4-4}\cline{7-9}
         &           &           & \multirow{4}{*}{$\checkmark$} &              &              &   6.9      &  13.4      & 10.1 \\ \cline{6-9}
         &           &           &              &              & $\checkmark$ &   6.3      &  13.1      & \textcolor{white}{0}9.7 \\ \cline{5-9}
         &           &           &              & \multirow{2}{*}{$\checkmark$} &              &   6.8      &  13.3      & 10.1 \\ \cline{6-9}
         &           &           &              &              & $\checkmark$ &   6.1      &  12.9      & \textcolor{white}{0}9.5 \\
\hline
\hline
 LSTM    &   Conf.   &  AdamW    &              &              &              &   7.6      &  13.9      & 10.7 \\
\hline
\end{tabular}                              
\label{tab:comp_lstm_conf}
\vspace{-2mm}
\end{table}

\vspace{-1.5mm}
\subsection{Experiments with conformers}
\label{sec:expconf}
\vspace{-1.5mm}
In the next set of experiments, we analyzed whether conformer or LSTM elements are more beneficial in the encoder and decoder modules of an attention based E2E model.
In Table~\ref{tab:comp_lstm_conf}, the conformer-LSTM encoder-decoder shows 0.4\% absolute gain over the LSTM-LSTM configuration.
I-vectors have inconsistent effects on conformer-LSTM E2E.
We also note that the conformer-encoder model ran significantly faster than the LSTM based one.
Using beam size of 4, the model achieved 0.12 real-time factor (RTF) without loss of accuracy on a single core of an Intel Xeon E5-2690v4.
After 8-bit integer quantization of the feed-forward and LSTM weights, we measured 0.08 RTF on a single core of an Intel Xeon Platinum 8280M CPU and 0.1\% WER degradation on Hub5'00.
Switching to a conformer based decoder turned out to be less efficient than a single head LSTM decoder, and the performance degraded.
For a more fair comparison, we also implemented scheduled sampling (SS) for the conformer decoder, and applied teacher forcing with 0.8 probability.
As can be seen, SS indeed improves the conformer decoder results (by 2\% relative), nevertheless the training time increases by more than 30\%.
Surprisingly, after decoding with external LM, the conformer-decoder results improve only 0.4\% absolute, compared to the 0.8-0.9\% gain we measure with conformer-LSTM model.
For the sake of completeness, we also ran experiments with LSTM-encoder conformer-decoder models.
Such model gives even worse results than the conformer-conformer architecture, and we measure 0.4\% absolute WER degradation.

\begin{table}
  \eightpt
  \centering
  \caption{Effect of external language model on conformer encoder-decoder model on Switchboard-300.}
  \vspace{-3.5mm}
  \begin{tabular}{|@{}c@{}|@{}c@{}|@{}c@{}|@{}c@{}|@{}c@{}|c|c|c|c|}
    \hline
 \multicolumn{4}{|c|}{LM}      &  PPL & \multicolumn{3}{c|}{WER [\%]}     \\ \hline
         & \multirow{2}{*}{\#param.} &  \multirow{2}{*}{xutt.} &  \multirow{2}{*}{\parbox{6mm}{prob.\\ratio}} & \multirow{2}{*}{CV}  & \multicolumn{3}{c|}{hub5'00} \\ \cline{6-8}
\hspace{12mm} & \hspace{11mm} & \hspace{7mm} & \hspace{8mm} & \hspace{8mm} &  swb    &    chm     & tot.   \\
\hline
\hline
         \multicolumn{5}{|c|}{n/a}             & 6.7      & 13.0       & 9.8  \\
\hline                                           
\hline
\multirow{4}{*}{LSTM}  &  \multirow{4}{*}{57M} &              &              & \multirow{2}{*}{52.9} & 5.9      & 12.2    & 9.0   \\ \cline{4-4}\cline{6-8}
         &           &              & $\checkmark$ &      & 5.8      & 11.8    & 8.8   \\ \cline{3-5}\cline{6-8}
         &           & \multirow{4}{*}{$\checkmark$} &              & \multirow{2}{*}{44.1} & 5.8      & 12.0    & 8.9   \\
         \cline{4-4}\cline{6-8}
         &           &              & \multirow{3}{*}{$\checkmark$} &      & 5.7      & 11.4    & 8.6   \\  \cline{1-2}\cline{5-8}
TrafoXL &  35M      &              &              & 39.3 & \bf 5.6      & \bf 11.2    & \bf 8.4   \\ \cline{1-2}\cline{5-8}
Conf.    &  42M      &              &              & 39.0 & \bf 5.6      & 11.3    & 8.5   \\
\hline
\hline
\multicolumn{5}{|l|}{LSTM\ +\ TrafoXL}                   & 5.5      & 11.2    & 8.4   \\ \cline{1-8}
\multicolumn{5}{|l|}{LSTM\ +\ TrafoXL + Conf.}           & \bf 5.4      & \bf 11.1    & \bf 8.3   \\
\hline
\end{tabular}                              
\label{tab:lm1}
\vspace{-5mm}
\end{table}

\begin{table*}
\eightpt
\centering
\caption{Overall results on Switchboard 300 and 2000 with single and combined end-to-end and cross-utterance language models.}
\vspace{-3.5mm}
\begin{tabular}{|@{}c@{}|@{}c@{}|@{}c@{}|@{}c@{}|@{}c@{}|@{}c@{}|@{}c@{}||@{}c@{}||c|c||c|@{}c@{}|@{}c@{}||c|c|}
\hline        
        & \multicolumn{4}{c|}{E2E}                & fea.ext      &  \multirow{2}{*}{ivec.} & \multirow{2}{*}{ext. LM} & \multicolumn{2}{c||}{hub5'00} & \multicolumn{3}{c||}{hub5'01} &     \multicolumn{2}{c|}{rt03} \\ \cline{2-5}\cline{9-15}
        &  id    & enc.  &   dec.     & \#parm. & step [ms]    &              &               &     swb   &   chm   &   swb   & \hspace{.6mm}swb2p3\hspace{.6mm} & \hspace{.6mm}swb2p4\hspace{.6mm} &   swb    & fsh  \\
\hline
\hline
\multirow{11}{*}{\rotatebox{90}{SWB-300}}
        &  1     & \multirow{5}{*}{LSTM} & \multirow{8}{*}{LSTM} &  57M    & \multirow{2}{*}{10} & \hspace{7mm} & \multirow{4}{*}{LSTM} & 6.0 & 12.0 & 6.6 & 8.8 & 12.8 & 13.6 & 7.5 \\ \cline{2-2}\cline{5-5}\cline{7-7}\cline{9-15}
        &  2     &       &            &  61M    & \hspace{14mm} & $\checkmark$ & \hspace{23mm}              & 5.8 & 12.1 & 6.6 & 8.6 & 12.8 & 14.0 & 7.7 \\ \cline{2-2}\cline{5-7}\cline{9-15}
        &  3     & \hspace{10mm} & \hspace{10mm} &  56M    & \multirow{3}{*}{2.5} &              &               & 6.0 & 12.0 & 6.7 & 8.6 & 12.5 & 13.2 & 7.7 \\ \cline{2-2}\cline{5-5}\cline{7-7}\cline{9-15}
        & \multirow{2}{*}{4} &       &            & \multirow{2}{*}{60M} &              & \multirow{2}{*}{$\checkmark$} &               & 5.9 & 11.5 & 6.5 & 8.5 & 11.6 & 13.0 & 7.2 \\ \cline{8-8}\cline{9-15}
        &        &       &            & \hspace{10mm} &              &              & LSTM+TrafoXL  & 5.7 & 11.3 & 6.1 & 8.3 & 11.3 & 12.6 & 7.0 \\  \cline{2-3}\cline{5-8}\cline{9-15}
        & \multirow{3}{*}{5} & \multirow{3}{*}{Conf.} &            & \multirow{3}{*}{68M}    & \multirow{3}{*}{10} &              &        -      & 6.7 & 13.0 & 7.1 & 9.2 & 13.7 & 15.7 & 9.1 \\ \cline{8-15}
        &        &       &            &         &              &              & LSTM          & 5.7 & 11.4 & 6.2 & 7.8 & 11.4 & 12.8 & 7.2 \\ \cline{8-15}
        &        &       &            &         &              &              & LSTM+TrafoXL  & 5.5 & 11.2 & 6.1 & 7.7 & 11.4 & 12.6 & 7.0 \\ \cline{2-2}\cline{3-15}
        & \multicolumn{6}{c||}{\multirow{2}{*}{1+3+5}}                                                              &     -         & 6.1 & 11.9 & 6.4 & 8.8 & 12.6 & 13.7 & 8.0 \\ \cline{8-15}
        & \multicolumn{6}{c||}{}                                                & \multirow{2}{*}{LSTM+TrafoXL}  & 5.1 & 10.1 & 5.5 & \bf 6.9 & 10.5 & \bf 10.7 & 6.1 \\ \cline{2-7}\cline{9-15}
        & \multicolumn{6}{c||}{2+4+5}                        &               & \bf 5.0 & \bf 10.0 & \bf 5.3 & 7.0 & \bf 10.4 & 10.8 & \bf 6.0 \\
\hline
\hline
\multirow{9}{*}{\hspace{1mm}\rotatebox{90}{SWB-2k}\hspace{1mm}} 
        &  6      & \multirow{2}{*}{LSTM} & \multirow{5}{*}{LSTM} &  661M   &     10      &              & \multirow{2}{*}{LSTM} & 4.6 & \textcolor{white}{0}7.8 & 5.2 & 6.1 & 10.0 &  \textcolor{white}{0}8.0 & 6.6 \\ \cline{2-2}\cline{5-6}\cline{9-15}
        &  7      &        &            &  663M   &    2.5      &              &               & 4.9 & \textcolor{white}{0}7.7 & 5.5 & 6.2 & \textcolor{white}{0}9.8 &  \textcolor{white}{0}8.4 & 6.7 \\ \cline{2-3}\cline{5-6}\cline{8-15}
        & \multirow{3}{*}{8} & \multirow{3}{*}{Conf.} &            & \multirow{3}{*}{\textcolor{white}{0}99M} & \multirow{5}{*}{10} &              &      -        & 4.8 & \textcolor{white}{0}8.0 & 5.2 & 6.4 & 10.3 &  \textcolor{white}{0}8.2 & 6.7 \\ \cline{8-15}
        &         &        &            &         &             &              & LSTM          & 4.7 & \textcolor{white}{0}7.6 & 5.0 & 6.1 & \textcolor{white}{0}9.7 &  \textcolor{white}{0}7.7 & \bf 5.9 \\ \cline{8-15}
        &         &        &            &         &             &              & LSTM+TrafoXL  & 4.6 & \textcolor{white}{0}7.6 & 4.9 & 6.2 & \textcolor{white}{0}9.5 &  \textcolor{white}{0}7.8 & \bf 5.9 \\ \cline{2-5}\cline{8-15}
        & \multirow{2}{*}{9} & \multirow{2}{*}{Conf.} & \multirow{2}{*}{Conf.} & \multirow{2}{*}{154M} &             &              &      -        & 4.7 & \textcolor{white}{0}7.6 & 5.0 & 6.1 & 10.0 & \textcolor{white}{0}8.0  & 7.0 \\ \cline{8-15}
        & \hspace{5mm} &        &            &         &             &              & \multirow{3}{*}{LSTM+TrafoXL} & 4.5 & \textcolor{white}{0}7.3 & 4.9 & 5.8 & \textcolor{white}{0}9.4 & \textcolor{white}{0}7.6  & 6.6 \\ \cline{2-7}\cline{9-15}
        & \multicolumn{6}{c||}{6+7}                                                                  &               & 4.5 & \textcolor{white}{0}7.2 & 4.9 & 5.7 & \textcolor{white}{0}9.2  &  \textcolor{white}{0}7.6 & 6.1 \\ \cline{2-7}\cline{9-15}
        & \multicolumn{6}{c||}{6+7+8+9}   &                                                                          & \bf 4.3 & \bf \textcolor{white}{0}6.8 & \bf 4.6 & \bf 5.5 & \bf \textcolor{white}{0}9.0 &  \bf \textcolor{white}{0}7.2 & \bf 5.9 \\
\hline
\end{tabular}                              
\label{tab:lm2}
\vspace{-4.5mm}
\end{table*}

\vspace{-1mm}
\subsection{Effect of decoding with external language model}
\vspace{-1mm}
Shallow fusion showed large improvement even with the best performing conformer model in Table~\ref{tab:comp_lstm_conf}.
Additional experiments were designed to investigate the effect of probability ratio fusion and self-attention LMs.
As can be seen in Table~\ref{tab:lm1}, incorporating the external LM through probability ratio fusion results in 0.2-0.3\% absolute WER improvement.
The gain is consistent even if cross-utterance external LM is used.
Further, transformer-XL LM gives additional 0.2\% reduction.
Conformer LM turned out to be not better than transformer-XL.
Since the probability ratio approach already corresponds to LM combination, we also ran recognition experiments with combined cross-utterance LSTM and attention based LMs.
As can be seen, slight improvement is measured over the best ``single'' LM result in Table~\ref{tab:lm1}.
Decoding with the best model runs at 0.6 RTF on a K80 GPU.

We also ran probability ratio LM fusion experiments with conformer decoders (results not presented), in order to get a better understanding of  the small effect of the external language model with such decoders we observed in Section~\ref{sec:expconf}.
Although the probability ratio approach improved the conformer decoder result more (0.4\% absolute WER improvement without i-vector), the conformer decoder still lagged significantly behind the LSTM decoder.
The gap remained roughly the same irrespective of the type and depth of the denominator LM.
We hypothesize that a deep multi-headed self-attention LM learns a significantly different mechanism than the decoder of a multi-headed ASR.
Thus, a simple model plug-in is not effective in recovering a score proportional to the emission likelihood $p(x_1^T|w_1^N)$.

\vspace{-1.5mm}
\subsection{Combination experiments}
\vspace{-1mm}
Besides decoding with the combination of self-attention, LSTM and probability ratio LMs, the application of multiple E2E models is also tested.
The detailed results of the single and combined systems are shown in the upper section of Table~\ref{tab:lm2}.
The superiority of the conformer over LSTM encoder is also confirmed on the Hub5'01 and RT03 sets.
The LSTM encoder trained on high temporal-resolution log-Mel and i-vector features is nevertheless fairly competitive, especially after decoding with combined LSTM+TrafoXL LM.
The combination of E2E models without external LM resulted in significant gains, in some cases over 10\% relative WER improvement over a very strong baseline.
E.g. WER on Hub5'00 CHM improved from 13.0\% to 11.9\%.
The i-vector has inconsistent effect on the combination.
Surprisingly, the effect of decoding with external language models is not mitigated by the combination of E2E models, and additional 16-30\% relative WER improvement is observed.
We also note that our best 300-hour results could even match system combination results on 2000 hours published only a few years ago \cite{capio}, which clearly indicates the great progress made in the recent years.

\vspace{-2mm}
\subsection{Experiments on Switchboard-2000}
\vspace{-.5mm}
Building similar systems on 2000 hours of speech data, we made the following observations.
Without retraining the UBM and total variability matrix, i-vectors showed inconsistent gain on the evaluation sets.
Scheduled sampling improved conformer decoder based  E2E model by 3\% relative on the larger dataset.
Not surprisingly, probability ratio fusion did not improve the results, because the E2E and LM were trained on the same dataset.
Moreover, switching from SGD to AdamW did not show consistent improvement with large scale models; the objective comparison can be made by contrasting the best LSTM results in \cite{Tuske2020} and the corresponding row in Table~\ref{tab:lm2}.

As results indicate in Table~\ref{tab:lm2}, a 100M-parameter conformer-encoder outperforms the large scale LSTM models.
On average, the best single-model performance is achieved by conformer-conformer model, but also note the discrepancy on the Fisher subset of RT03.
Overall, system and language model combination improves the state-of-the-art by 8-14\% relative, compared to the previous best numbers reported in \cite{Tuske2020}.
The improvement is also confirmed on RT02 and RT04 sets, where the best combination achieves 6.3\% and 5.2\% WER.
The performance of the best system combination on SWB part of Hub5'00 (4.3\%) is clearly below the human error rate (5.1\%) measured in \cite{Saon2017}.
This can be attributed to fact that most of the speakers appear in the training data, decoder hyperparameters are optimized on Hub5'00, and the human error rate might also have been overestimated.
We note that the 4-E2E combination with multiple LMs can run on a modern V100 GPU at 0.4 RTF.
Considering that human performance, and thus transcription error rate, is at 6.8\%, 6.0\%, 4.5\%, 4.7\% WER on the CHM subset of Hub5'00 and on the RT0\{2,3,4\} sets \cite{Saon2017}, it is surprising to see how close we can actually get with 2000 or even only 300 hours of data using fairly simple and general attention models.

\vspace{-1mm}
\section{Conclusions}
\vspace{-.5mm}
\label{sec:conc}
\renewcommand{\baselinestretch}{0.946}\normalsize
We investigated a set of techniques on top of our best recipe for the Switchboard English speech recognition benchmark.
We showed that complementary systems with similar performances can be developed using different input features or diverse encoder, decoder and language model structures.
Using a more advanced optimizer and score combination techniques, our final Switchboard-300 system achieved 5.0\% and 10.0\% WER on the SWB and CHM subsets of Hub5'00.
Joint decoding with such models trained on Switchboard-2000 achieves a new state of the art: the flexible attention models are able to reach the limit of many of the standard evaluation sets.
Future work could focus on how to achieve such performance with a limited amount of transcriptions.

\newpage
\ninept
\bibliographystyle{IEEEtran}
\bibliography{mybib}

\end{document}